# MAIES: A Tool for DNA Mixture Analysis

**Robert G. Cowell**
Faculty of Actuarial Science and Insurance
Sir John Cass Business School
106 Bunhill Row
London EC1Y 8TZ, UK.

**Steffen L. Lauritzen**
Department of Statistics
University of Oxford
1 South Parks Road
Oxford OX1 3TG, U.K.

**Julia Mortera**
Dipartimento di Economia
Università Roma Tre
Via Ostiense, 139
00154 Roma, Italy.

## Abstract

We describe an expert system, MAIES, under development for analysing forensic identification problems involving DNA mixture traces using quantitative peak area information. Peak area information is represented by conditional Gaussian distributions, and inference based on exact junction tree propagation ascertains whether individuals, whose profiles have been measured, have contributed to the mixture. The system can also be used to predict DNA profiles of unknown contributors by separating the mixture into its individual components. The use of the system is illustrated with an application to a real world example. The system implements a novel MAP (*maximum a posteriori*) search algorithm that is briefly described.

## 1 Introduction

Probabilistic expert systems (PES) for evaluating DNA evidence were introduced in [1]. This paper is concerned with describing the current status of a computer software system called MAIES (Mixture Analysis in Expert Systems) that analyses *mixed traces* where several individuals may have contributed to a DNA sample left at a scene of crime. In [2] it was shown how to construct a PES using information about which alleles were present in the mixture, and we refer to this article for a general description of the problem and for genetic background information. (A brief summary to genetic terminology is given in Appendix A.)

The results of a DNA analysis are usually represented as an *electropherogram* (EPG) measuring responses in *relative fluorescence units* (RFU) and the alleles in the mixture correspond to peaks with a given height and area around each allele, see Figure 1. The band intensity around each allele in the relative fluorescence units represented, for example, through their *peak areas*, contains important information about the composition of the mixture.

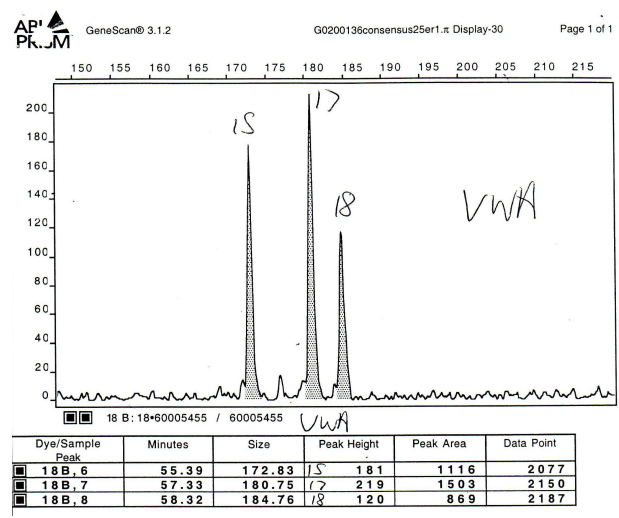

Figure 1: An electropherogram (EPG) of marker VWA from a mixture. Peaks represent alleles at 15, 17 and 18 and the areas and height of the peaks express the quantities of each. Since the peak around allelic position 17 is the highest this indicates that the 17 allele is likely to be a homozygote or a shared allele between two heterozygotes. This image is supplied courtesy of LGC Limited, 2004.

The main focus of the present paper is to describe the current status of a computer package called MAIES, which automatically builds Bayesian network models for mixture traces based on conditional Gaussian distributions [3] for the peak areas, given the composition of the true DNA mixtures. Currently the program only considers a DNA mixture from exactly two contributors, which seems to be the most common scenario in forensic casework [4], and the program ignores other important complications such as stutter, dropout alleles, etc.

We distinguish two types of calculations that MAIES can perform. One type is *evidential calculation*, in which a *suspect* with known genotype is held and we want to determine the likelihood ratio for the hypothesis that the suspect has contributed to the mixture vs. the hypothesis that the contributor is a randomly chosen individual. We distinguish two cases: the other contributor could be a *victim* with a known genotype or a *contaminator* with an unknown genotype, possibly without a direct relation to the crime. This could be a laboratory contamination or any other source of contamination from an unknown contributor. The other type calculation that MAIES can perform is the *separation of profiles,* i.e. identifying the genotype of each of the possibly unknown contributors to the mixture, the evidential calculation playing a secondary role. Both types of calculation are illustrated in §5.

Previous related work includes that of [5] and [6] who respectively developed numerical methods known as *Linear Mixture Analysis* (LMA) and *Least Square Deconvolution* (LSD) for separating mixture profiles using peak area information. Both methods are based on least squares heuristics that assume the mixture proportion of the contributors' DNA in the sample is constant across markers. A computer program has been written [7] for estimating the proportion of the individual contributions in two-person mixtures and to rank the genotype combinations based on minimizing a residual sum of squares. More recently, [8] describes PENDULUM, a computer package to automate guidelines in [9] and [7]. None of the methods described above utilizing peak area information are probabilistic in nature, nor do they use information about allele frequency. In contrast, the methodology proposed in [10] combines a model using the gene frequencies with a model describing variability in scaled peak areas to calculate likelihood ratios and study their sensitivity to assumptions about the mixture proportions.

The plan of the rest of the paper is as follows. In the following section we describe the mathematical assumptions underlying the Bayesian networks that MAIES generates for analysing two-person DNA mixtures. We then describe the components that MAIES uses to build up the networks. This is followed by a description of a simple MAP search algorithm, implemented in MAIES for separation of profiles. We then illustrate the use of MAIES on a real life example, and then summarize future work required to make MAIES into a tool for routine casework.

## 2 The mathematical model

Our PES is a probabilistic model for relating the pre-amplification and post-amplification relative amounts of DNA in a mixture sample. The model is idealized in that it ignores complicating artefacts such as stutter, drop-out alleles and so on, and assumes that the mixture is made up of DNA from two people, who we refer to as p1 and p2. Typically, prior to amplification in a laboratory, a DNA mixture sample will contain an unknown number of cells from p1 and a further unknown number of cells from p2. Hence there is an unknown common fraction, or proportion, across the markers of the amount of DNA from p1, that we denote by $\theta$. In an ideal amplification apparatus, during each amplification cycle the proportion of alleles of each allelic type would be preserved without error. We model departures from this ideal as random variation using the Gaussian distributions, and we introduce an additional variance term to represent other measurement error, represented by $\omega^2$.

The post-amplification proportions of alleles for each marker are represented in the peak area information, which we include in the analysis through the *relative peak weight*. The (absolute) *peak weight* $w_a$ of an allele with *repeat number* $a$ is defined by scaling the peak area with the repeat number as $w_a = a\alpha_a$, where $\alpha_a$ is the peak area around allele $a$. Multiplying the area with the repeat number is a crude way of correcting for the fact that alleles with a high repeat number tend to be less amplified than alleles with a low repeat number. For issues concerning heterozygous imbalance see [11].

We further assume that

- The peak weight for an allele is approximately proportional to the amount of DNA of that allelic type;
- The peak weight for an allele possessed by both contributors is the sum of the corresponding weights for the two contributors.

To avoid the arbitrariness in scaling used to measure the areas, we consider the observed *relative peak weight* $r_a$, obtained by scaling with the total peak weight as

$$r_a = w_a/w_+, \quad w_+ = \sum_a w_a,$$

so that then $\sum_a r_a = 1$.

For the relative peak weight, denoted by the random variable $R_a$, we assume a Gaussian error distribution

$$R_a \sim \mathcal{N}(\mu_a, \tau_a^2), \quad \mu_a = \{\theta n_a^{(1)} + (1-\theta)n_a^{(2)}\}/2, \quad (1)$$

where $\theta$ is the proportion, or fraction, of DNA in the mixture originating from the first contributor, $n_a^{(i)}$ is the number of alleles with repeat number $a$ possessed by person $i$. Note that $n_a^{(i)} \in \{0,1,2\}$ and hence $\mu_a \in [0,1]$.

We assume an error variance for $\tau_a^2$ of the form

$$\tau_a^2 = \sigma^2 \mu_a(1-\mu_a) + \omega^2 \qquad (2)$$

where $\sigma^2$ and $\omega^2$ are variance factors for the contributions to the variation from the amplification and measurement processes.[1] Note that if $\mu_a = 0$ then $\tau_a^2 = \omega^2$ and $R_a \sim \mathcal{N}(0, \omega^2)$. The interpretation of this is that if there are no alleles of type $a$ in the mixture prior to amplification, then any detected post-amplification can be ascribed to measurement error. Similarly if $\mu_a = 1$, which means that for the given marker all alleles are of type $a$ in the mixture before amplification, then $R_a \sim \mathcal{N}(1, \omega^2)$, that is post amplification all alleles for the given marker are of type $a$, up to measurement error.

However the Bayesian networks that MAIES constructs uses the variance structure

$$\tau_a^2 = \sigma^2 \mu_a + \omega^2. \qquad (3)$$

The reason is that we need to consider the correlation between weights due to the fact that they must add up to unity. It turns out, perhaps surprisingly, that the likelihood obtained using (3) when conditioned on $\sum_a r_a = 1$ has precisely the form as would be obtained using the likelihood based on (2) used in our model if we ignore measurement error by setting $\omega^2 = 0$, and the likelihoods are very close numerically for small $\omega^2$. This constraint $\sum_a r_a = 1$ is imposed when we enter the complete set of observed peak weights as evidence in our networks. The proof, too long for this paper, may be found in [12].

In our example in §5 we used $\sigma^2 = 0.01$ and $\omega^2 = 0.001$, corresponding approximately to a standard deviation for the observed relative weight of about $\sqrt{0.01/4 + 0.001} = 0.06$ for $\mu_a = 0.5$ substituted into (2). These parameter values imply that when amplifying DNA from one heterozygous individual (for which $\mu_a = 0.5$), an $r_a$ value at two standard deviations from the mean would give a value of $0.38/0.62 = 0.61$ for the ratio of the minor to the major peak area; this is about the limit of variability in peak imbalance that has been reported in the literature [13], and suggests that our chosen parameter values are perhaps conservative.

In general the variance factors may depend on the marker and on the amount of DNA analysed, but for simplicity we use the values above. (Our PES model is robust to small changes in these parameter estimates.)

Finally, we assume known gene frequencies of single STR alleles; in particular we use those reported in [14] for U.S. Caucasians for the analysis in §5 of data taken from [9][2]

## 3 Maies

The basic form of our Bayesian network models is fairly straightforward, but the networks can grow large when modelling the ten or so markers typical in a mixture problem. (In the example in §5 the network has 237 nodes.) One way to manage this complexity is to use object-oriented Bayesian network software, as we describe in detail in [12]. Here we describe MAIES, a purpose built program that, after entering peak area information and available genetic information (if available) about the potential contributors, automatically constructs a single conditional-Gaussian Bayesian network on which the probability calculations are performed. MAIES implements the local propagation scheme of [15]. Peak areas are automatically converted to normalized weights and entered as evidence in the relevant nodes by the program.

An example of a network generated for a single marker with two alleles observed in the mixture is shown in Figure 2. The figure illustrates the repetitive modular structure that makes it possible for MAIES to create the much larger Bayesian networks required to analyse mixtures on several markers. We now describe these various structures and how they interrelate, working from top to bottom in Figure 2.

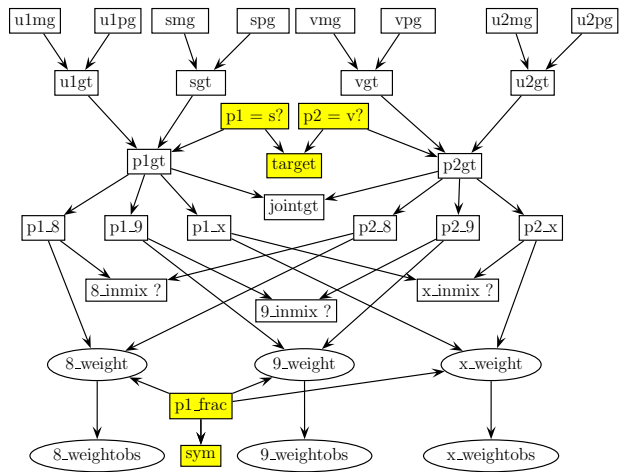

Figure 2: The structure of a Bayesian network generated by MAIES for a single marker, in which two allele peaks (8 and 9) were observed.

---

[1]The first term in the variance structure in (2) can be seen as a second order approximation to a more sophisticated model based on gamma distributions for the absolute scaled peak weights to be discussed elsewhere.

[2]This dataset has an observed allele 36 of the marker D21. As none of the 302 subjects in [14] had this allele, we chose to use 1/604=0.00166 as its frequency.

## 3.1 Top level people

MAIES currently models mixtures only for DNA from two individuals. Thus it sets up nodes for four individuals who are paired up, prefixed by `s` (for suspect), `v` (for victim), and `u1` and `u2` representing two unspecified persons from the population. Corresponding to each of these individuals is a triple of nodes representing their genotype (gt) on the marker, and the individuals' paternal (pg) and maternal (mg) genes. The probability tables associated with the maternal and paternal genes contain the allele frequencies of the observed alleles, whilst the conditional probability table associated with the genotype node is the logical combination of the maternal and paternal gene.

## 3.2 Actual contributors to the mixture

The genotypes on the marker of the two individuals p1 and p2 whose DNA is in the mixture are the nodes labelled `p1gt` and `p2gt`. Node `p1gt` has incoming arrows from nodes `u1gt`, `sgt` and a (yes,no) valued binary node labelled `p1 = s?`. The function of this latter node is to set the genotype of node `p1gt` to be that of `sgt` if `p1 = s?` takes the value yes, otherwise to set the genotype of node `p1gt` to be that of `u1gt`. An equivalent relationship holds between the genotype nodes `p2gt`, `vgt`, `u2gt` and `p2 = v?`. Uniform priors are placed on the nodes `p1 = s?` and `p2 = v?`.

The node labelled `target` represents the four possible combinations of values of the two nodes `p1 = s?` and `p2 = v?`, with a conditional probability table of zeros and ones representing the logical identities. The marginal posterior distribution of this node is used to calculate likelihood ratios in evidential calculations.

The network also has a node representing the joint genotypes of individuals p1 and p2, which is labelled `jointgt`, with incoming arrows from `p1gt` and `p2gt`; the (quite big) conditional probability table associated with this node has entries that are either of zero or one. The most likely configuration of the marginal distribution in all joint genotype nodes across all markers is required for separating the mixture, and is found using the MAP search algorithm described in § 4.

## 3.3 Allele counting nodes

On the level below the genotype nodes for p1 and p2 is a set of nodes representing the number of alleles (taking the value of 0, 1 or 2) of a certain type in each individual. Thus, for example, the node `p1_8` counts the number of alleles of repeat number 8 in the genotype of individual p1 for the given marker: this value only depends upon the genotype of the individual p1 and hence there is an arrow from `p1gt` to `p1_8`.

These nodes model the $n_a^{(i)}$ variables introduced in (1).

## 3.4 Repeat number nodes

On the level below the allele counting nodes are the repeat number nodes, labelled `8_inmix?`, `9_inmix?` and `x_inmix?`. These are (yes,no) binary valued nodes representing whether or not the particular alleles are present in the mixture: thus for example allele 8 is present in the mixture if either of the allele counting nodes `p1_8` or `p2_8` takes a non-zero value. For the node `x_inmix?` the `x` refers to all of the alleles in the marker that are not observed. When using repeat number information as evidence the repeat number nodes present in the mixture will be given the value yes and `x_inmix?` will be given the value no.

## 3.5 True and observed weight nodes

These nodes are represented by the elliptical shapes. The nodes `8_weight`, `9_weight` and `x_weight` represent the true relative peak weights $r_8$, $r_9$ and $r_x$ respectively of the alleles 8, 9 and x in the amplified DNA sample. Each true-weight node is given a conditional-Gaussian distribution as in (1), where the fraction $\theta$ of DNA from p1 in the mixture is modelled in the network by a discrete distribution in the node labelled `p1_frac`. The variance is taken to be $\sigma^2 \mu_a$. The nodes `8_weightobs`, `9_weightobs` and `x_weightobs` represent the measured weights. The observed weight is given a conditional-Gaussian distribution with mean the true weight, and measurement variance $\omega^2$, hence leading to the variance (3).

When using peak area information as evidence the nodes representing the observed weights will have their values set to the relative peak weights. The `sym` node is only used for separating a mixture of two unknown contributors, to break the symmetry between p1 and p2 (see § 5.2).

## 3.6 Networks with more than one marker

The network displayed in Figure 2 generated by MAIES is for a single marker; for mixture problems involving several markers the structure is similar but more complex because the number of nodes grows with the number of markers. In such a network the nodes shaded in Figure 2 occur only once. The unshaded nodes are replicated once for each marker, with each node having text in their labels to identify the marker that the allele or genotype nodes refer to. There will also be extra repeat number, allele counting and allele weight nodes in each marker having more than two observed alleles in the mixture, extending the pattern for the one-marker network in the obvious manner.

## 4 A simple MAP search algorithm

It is well known that the most likely configuration of a set of discrete variables is not necessarily the same as found by picking the most likely states in the individual marginals of the variables (see for example [16]). The basis of the MAP search algorithm in MAIES is to assume that this is close.

Specifically, after entering and propagating evidence, one finds the individual marginals of the set of MAP variables $M$ of interest. There are then two variants of the MAP algorithm, *batch* and *sequential*.

In the batch variant, one finds for some reasonable number $n$ (say $n = 5000$) the top $n$ most likely configurations of the joint probability given by the product of the individual marginals of MAP variables. This is done by constructing a disconnected graph in the MAP variables, and using the efficient algorithm of [17]. These configurations are stored in an list $(c_1, c_2, \ldots, c_n)$ ordered by decreasing probability according to the independence graph. Now returning to the original Bayesian network, one propagates the available evidence $\mathcal{E}$ and finds the normalization constant $P(\mathcal{E})$ and stores this in $rp$, say (short for "remainder probability"). One then processes the $(c_1, c_2, \ldots, c_n)$ configurations as additional evidence in the original Bayesian network and finds from the normalization constant each of their probabilities $P(c_i, \mathcal{E})$. After processing each configuration, one keeps track of the highest probability configuration found and its probability, $bp$. One also subtracts $p(c_i, \mathcal{E})$ from $rp$, so that if it ever happens $bp > rp$ then the MAP has been found. If one stores all of the probabilities $p(c_i, \mathcal{E}), i = 1, \ldots, k$ for all of the configurations that have been processed, then perhaps the second, third etc. most likely configurations may be identified if their probabilities exceed $P(\mathcal{E}) - \sum_{i=1}^{k} P(c_i, \mathcal{E})$. The sequential variant proceeds similarly, the difference is that the candidates $c_1, c_2, \ldots$, are generated one at a time as required. The following is pseudo-code for the sequential variant for finding the MAP.

- Initialize: $i = j = 1$, $bp = 0$, and $rp = P(\mathcal{E})$.
- While $bp < rp$ do:
  - Find $c_i$ and $P(c_i, \mathcal{E})$;
  - If $p(c_i, \mathcal{E}) > bp$ set $bp = P(c_i, \mathcal{E})$ and $j = i$;
  - Set $rp := rp - P(c_i, \mathcal{E})$ and $i := i + 1$;
- $c_j$ is the MAP configuration.

For purely discrete networks, this algorithm does not appear to be as efficient as that described in [16]. However it is neither clear that the latter can be applied to finding the MAP of a set of discrete variables in a conditional Gaussian network, nor that it could identify the second, third, fourth etc., most likely configurations.

## 5 A criminal case example

Our example is taken from Appendix B of [9] and illustrates the use of the *amelogenin* marker in the analysis of DNA mixtures when the individual contributors are of opposite sex.

Peak area analysis of the amelogenin marker in DNA recovered from a condom used in a rape attack indicated an approximate 2:1 ratio for the amount of female to male DNA contributing to the mixture. Peak area information was available on six other markers, the information is shown in Table 1; we shall refer to this as the *Clayton* data. (Further examples are illustrated in [12] and [18].)

Table 1: *Clayton* data of [9] showing mixture composition, peak areas and relative weights together with the DNA profiles of both victim (v) and suspect (s). For the marker D21 the allele designation in brackets is as given in [9] using the labelling convention of [19]

| Marker | Alleles | Peak area | Relative weight | s | v |
|---|---|---|---|---|---|
| Amelogenin | X | 1277 | 0.8298 | X | X |
| | Y | 262 | 0.1702 | Y | |
| D8 | 13 | 3234 | 0.6372 | | 13 |
| | 14 | 752 | 0.1596 | 14 | |
| | 15 | 894 | 0.2032 | 15 | |
| D18 | 14 | 1339 | 0.1462 | 14 | |
| | 15 | 1465 | 0.1714 | 15 | |
| | 16 | 2895 | 0.3612 | | 16 |
| | 18 | 2288 | 0.3212 | | 18 |
| D21 | 28 (61) | 373 | 0.1719 | 28 | |
| | 30 (65) | 590 | 0.2913 | | 30 |
| | 32.2 (70) | 615 | 0.3259 | | 32.2 |
| | 36 (77) | 356 | 0.2109 | 36 | |
| FGA | 22 | 534 | 0.1547 | 22 | |
| | 23 | 2792 | 0.8453 | 23 | 23 |
| THO | 5 | 5735 | 0.2756 | | 5 |
| | 7 | 10769 | 0.7244 | 7 | 7 |
| VWA | 15 | 1247 | 0.1633 | 15 | |
| | 16 | 1193 | 0.1667 | 16 | |
| | 17 | 2279 | 0.3383 | | 17 |
| | 19 | 2000 | 0.3318 | | 19 |

### 5.1 Evidential calculation

One possible use of the system to the *Clayton* data would be to compare the two hypotheses:

- $H_0$: the suspect and victim both contributed to the mixture
- $H_1$: the victim and an unknown contributor contributed to the mixture

In a courtroom setting, the null hypothesis, $H_0$, would be a prosecution's case, whilst $H_1$ would represent the defence's case. (It is standard procedure in court for likelihood ratios of these hypotheses to be reported.)

To do this calculation in MAIES, evidence is entered in the observed relative peak area nodes, the repeat number nodes, and information on the suspect and victim genotypes. After propagating the evidence the marginal on the `target` is examined. This has the following values (taken from MAIES):

$$\begin{array}{ll} \text{u1 \& u2} & 4.2701211814389 \times 10^{-21}, \\ \text{v and u} & 4.1040333719867 \times 10^{-11}, \\ \text{s and u} & 3.660791624072 \times 10^{-11}, \\ \text{s and v} & 0.99999999992235. \end{array}$$

From this the likelihood ratio of $H_0$ to $H_1$ is calculated to be $P(s \text{ and } v \,|\, \mathcal{E})/P(s \text{ and } u \,|\, \mathcal{E}) = 2.73 \times 10^{10}$, where $\mathcal{E}$ denotes the complete set of evidence. (Note that because we have placed uniform priors on the nodes `p1 = s?` and `p2 = v?`, then $P(\mathcal{E} \,|\, s \text{ and } v)/P(\mathcal{E} \,|\, s \text{ and } u) = P(s \text{ and } v \,|\, \mathcal{E})/P(s \text{ and } u \,|\, \mathcal{E})$.)

It may be that only DNA from a suspect is available, but not from a victim. In such a situation we could use MAIES to compare the following two hypotheses:

- $H_0$: the suspect and an unknown contributor contributed to the mixture
- $H_1$: two unknown contributors contributed to the mixture

Again in a courtroom setting these could represent prosecution and defence cases respectively. The calculation proceeds as before, but with the victim profile omitted from the evidence. This time the marginal on the target node is given by

$$\begin{array}{ll} \text{u1 \& u2} & 5.8322374221768 \times 10^{-11}, \\ \text{v and u} & 5.8322374221768 \times 10^{-11}, \\ \text{s and u} & 0.49999999994168, \\ \text{s and v} & 0.49999999994168, \end{array}$$

and the likelihood ratio of $H_0$ vs. $H_1$ is given by $P(s \text{ and } u \,|\, \mathcal{E})/P(u1 \text{ and } u2 \,|\, \mathcal{E}) = 8.57 \times 10^9$.

## 5.2 Mixture separation calculations

The other type of calculation that may be performed with MAIES is that of separating a two-person mixture into genotypes of the contributors. The output from such a decomposition could be used to find a match in a DNA database search. For such a search it is useful to have not just the most likely combination of genotypes of the two contributors to the mixture, but other less likely but also plausible combinations. MAIES achieves this with the MAP search algorithm described in § 4.

For separating a mixture, we may or may not have genetic information about one of the contributors. For our rape example, suppose that we have the genotype of the victim. Then using MAIES we may enter as evidence the victim's genotype, the relative peak areas and the repeat number information. We also select the value `yes` in the `p2 = v?` node. We then select the set of joint genotype nodes, and perform the MAP search. MAIES returns two configuration, the most likely having posterior probability 0.997594, with the genotype p1 matching our suspect profile in Table 1. The second most likely combination has a posterior probability of 0.00239796, and differs from the true profile in the marker FGA where a homozygous $(22, 22)$ genotype is predicted. All remaining possible genotype combinations have a total probability mass of less than $8 \times 10^{-6}$.

The second possibility is that no genotypic information is available on either contributor to the mixture. To do this calculation, evidence on the observed relative peak areas and the repeat number information is entered. To overcome the symmetry in the network between p1 and p2, we enter evidence on the `sym` node that `p1_frac` is $\geq 0.5$. Then, selecting the joint genotype nodes as before, we perform a MAP search. The result is shown in Table 2. All markers are correctly identified. Note in particular that the genotypes for the marker THO are identified correctly. In [9] this was only possible to do so after the victim's profile was taken into account.

Table 2: Most likely genotype combination of both contributors for *Clayton* data. The victim (here p1) and male suspect (p2) is correctly identified on every marker. The final column indicates the marginal probabilities for the genotype pairs on individual markers, with the figure in parenthesis the product of these marginals.

| Marker | Genotype p1 | Genotype p2 | Posterior probability |
|---|---|---|---|
| Amelogenin | X X | X Y | 0.983115 |
| D8 | 13 13 | 14 15 | 0.903013 |
| D18 | 16 18 | 14 15 | 0.993166 |
| D21 | 30 32.2 | 28 36 | 0.945235 |
| FGA | 23 23 | 22 23 | 0.989090 |
| THO | 5 7 | 7 7 | 0.845031 |
| VWA | 17 19 | 15 16 | 0.992738 |
| joint | 0.701988 | | (0.691517) |

For this example, the MAIES MAP search algorithm identifies the next three most likely combinations of genotypes, these have probabilities of 0.120049, 0.0583912 and 0.0227133 respectively. (The network has 183 discrete nodes and 54 continuous nodes. The total state space of the 7 joint genotype nodes is approximately $5.9 \times 10^{12}$. Identifying the 15 most likely combinations of these nodes took approximately 38 seconds on a 1.6GHz laptop with 256Kb memory.)

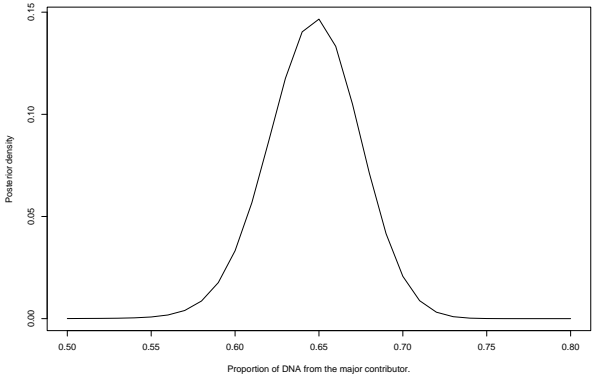

Figure 3: Posterior distribution of mixture proportion from *Clayton* data using no genotypic information.

Finally for this example, Figure 3 shows the posterior distribution of the mixture proportion; the peak at around 0.65 corresponds to a mixture ratio of 1.86:1, in line with the approximate 2:1 estimated in [9].

## 6 Discussion

We have described a software system, MAIES, for analysing DNA mixtures using peak area information, yielding a coherent way of predicting genotypes of unknown contributors and assessing evidence for particular individuals having contributed to the mixture, and applied it to a real life example. A simple MAP search algorithm allows a set of most plausible genotypes to be generated, perhaps for use in a DNA database search for a suspect.

There are a number of issues that would need addressing before the system could be used in routine analysis of casework, for example, complications such as more than two potential contributors, multiple traces, indirect genotypic evidence, stutter, etc. In addition, preliminary investigations seem to indicate that the variance factor depends critically on the total *amount* of DNA available for analysis. As this necessarily is varying from case to case, a calibration study should be performed to take this properly into account. Methods for diagnostic checking and validation of the model should be developed based upon comparing observed weights to those predicted when genotypes are assumed correct. Such methods could also be useful for calibrating the variance parameters $\sigma^2$ and $\omega^2$. We are pursuing ways that this could be accomplished using an EM estimation algorithm. (Bayesian methods for estimating the variance parameters could also be developed.) Nevertheless, despite these many issues, we feel that the present framework provides a sound foundation in which these and other matters can be be addressed and incorporated into MAIES.

### Acknowledgements

This research was supported by a Research Interchange Grant from the Leverhulme Trust. We are indebted to participants in the above grant and to Sue Pope and Niels Morling for constructive discussions. We thank Caryn Saunders for supplying the EPG image used in Figure 1.

## A  Some genetic background

Every cell nucleus of a person contains 46 chromosomes, which can be grouped into 22 distinguishable pairs of so called *autosomal* chromosomes, and one pair of sex-linked chromosomes (in men the latter is denoted by $XY$, in women by $XX$).

Each chromosome consists of a sequence of large molecules called *nucleotides*, there being four nucleotides in all (these are adenine, cytosine, guanine and thymine). Genes are particular subsequences in the paired chromosomes; any specific gene is associated with a specific chromosome at a particular position called a *locus* (hence there are two genes at a locus of a chromosome pair). A gene used in forensic identification at a particular locus is called a *marker*. *Short tandem repeat* (STR) markers are a particular type of marker favoured by forensic scientists in identifying people using nuclear DNA. They consist of repeated sequences of molecules called *base-pairs*, a single sequence consists of a small number of base-pairs. A marker may have one of a (finite) number of alternative molecular compositions, called *alleles*. The *allele number* or *repeat number* of an allele counts how many copies of a repeated sequence it consists of. Thus for example, allele 8 of the VWA marker system is an allele with eight repeats of a base-pair sequence. Allele numbers that are not integers indicate that one of the repeated sequences is not complete. For example, allele 9.3 of the VWA system contains nine repeats of a base-pair sequence plus a subsequence of three base pairs. The number of possible alleles will vary according to the particular marker. An unordered pair of alleles is associated with each marker. One of these is inherited from the father, and is called the *paternal gene*, the other is inherited from the mother and is called the *maternal gene*. This pair is called the *genotype*. If the pair of alleles has the same repeat number, then the genotype is called *homozygous*, otherwise it is called *heterozygous*.

The various alleles of a particular marker do not occur with equal frequency among the members of a population. By making DNA measurements of many individuals in a population the relative frequencies of alleles in a population may be estimated.